\documentclass{article}

\PassOptionsToPackage{numbers}{natbib}
\usepackage[nonatbib,preprint]{neurips_2024}

\RequirePackage{appendix}
\usepackage[
backend=biber,
maxbibnames=99
]{biblatex}
\usepackage[T1]{fontenc}    
\usepackage{hyperref}       
\usepackage{url}            
\usepackage{booktabs}       
\usepackage{amsfonts}       
\usepackage{nicefrac}       
\usepackage{microtype}      
\usepackage{xcolor}         
\usepackage{graphicx} 
\usepackage{subcaption}
\usepackage{amsmath}
\usepackage[ruled,linesnumbered]{algorithm2e}

\newcommand\numberthis{\addtocounter{equation}{1}\tag{\theequation}}

\graphicspath{ {./graph/} }

\title{Incremental Structure Discovery of Classification via Sequential Monte Carlo}

\author{%
  Changze Huang, Di Wang\thanks{Corresponding Author}\\
  Key Laboratory of High Confidence Software Technologies (Peking University), Ministry of Education\\
  School of Computer Science, Peking University, Beijing, China\\
  \texttt{hcz@stu.pku.edu.cn, wangdi95@pku.edu.cn} \\
}

\begin{document}

\maketitle

\begin{abstract}
Gaussian Processes (GPs) provide a powerful framework for making predictions and understanding uncertainty for classification with kernels and Bayesian non-parametric learning. Building such models typically requires strong prior knowledge to define preselect kernels, which could be ineffective for online applications of classification that sequentially process data because features of data may shift during the process. To alleviate the requirement of prior knowledge used in GPs and learn new features from data that arrive successively, this paper presents a novel method to automatically discover models of classification on complex data with little prior knowledge. Our method adapts a recently proposed technique for GP-based time-series structure discovery, which integrates GPs and Sequential Monte Carlo (SMC). We extend the technique to handle extra latent variables in GP classification, such that our method can effectively and adaptively learn a-priori unknown structures of classification from continuous input. In addition, our method adapts new batch of data with updated structures of models. Our experiments show that our method is able to automatically incorporate various features of kernels on synthesized data and real-world data for classification. In the experiments of real-world data, our method outperforms various classification methods on both online and offline setting achieving a 10\% accuracy improvement on one benchmark. 
\end{abstract}

\section{Introduction}

Classification is a fundamental problem in machine learning research methods \cite{RN61, RN63, 726791, 10.1007/978-3-540-30115-8_22}. Many outstanding solutions have been proposed with different perspective, including neural networks \cite{RN60}, tree model \cite{10.1145/2939672.2939785}, kernel method \cite{708428}, etc. These distinct methodologies offer unique advantages across various problem domains, necessitating a profound understanding of both the problem domain and algorithmic intricacies  to discern and implement the most suitable solution. 

The design of classification methods widely faces problems of insufficient prior knowledge and variation of data patterns. In order to automatically select a suitable method for a specific domain, meta-learning \cite{SHAO2023211, reif2012meta, jeong2020ood} evaluates various candidate methods and chooses the method that, in its opinion, fits the problem best. However, building such candidate methods would also require expertise and  it is possible that the pattern of a classification domain may shift across time or batches of inputs, especially in an online setting. Such a problem can be referred as \textit{incremental learning}. 

In this work, we study incremental learning for Gaussian Process (GP) classification with automatic selection of GP kernels. Our method follows the methodology of Bayesian inference. We define an assumption as a \emph{prior} distribution that describes the distribution of GP kernels, including both the kernel structures and parameters. Then, the \emph{evidence} or \emph{likelihood} is obtained by observing given data points to perform conditioning. By the Bayes' rule, one obtain the \emph{posterior} distribution of GP kernels, in a manner that is consistent with the prior distribution and the evidence. 

The reason that we choose Bayesian learning to address the problem is its ability to describe distribution of kernels, quantify uncertainty in predictions and adapt to new data by updating the \emph{posterior} distribution. Describing distribution of kernels is fundamental for automatic selection for kernels because it provides a probabilistic framework to evaluate and compare different kernels based on their suitability for the data. With quantification of uncertainty and adjustment of the \emph{posterior} distribution, our method is flexible and robust in dynamic and evolving datasets.

AutoGP \cite{saad2023sequential} is a recently proposed framework of automatic selection of GP kernels for regression based on Bayesian learning. It proposes a structure learning algorithm that combines sequential Monte Carlo (SMC) and Markov Chain Monte Carlo (MCMC) methods for efficient posterior inference.
For GP regression, the posterior distribution is usually Gaussian so one can use analytical closed form solutions; however,
for GP classification, one usually needs to handle non-Gaussian posterior distributions, due to an extra mapping from GP-produced logits to actual classification probabilities.   

\paragraph{Contribution}
We propose a novel method for automatically selecting Gaussian kernels for models of classification and incrementally learning from evolving datasets. We expand the framework of AutoGP to the domain of binary classification by automatic selection of kernel, which reduces requirement of prior knowledge. Our method applies Sequential Monte Carlo with online kernel adaption of classification to address the problem of pattern shifting during online learning. Our experiment shows that our method can automatically discover different kernel structures in different datasets and outperform  various classification methods on real-world datasets.

\section{Gaussian Processes Classification Models}
\label{automatic}

In this section, we formulate a family of Gaussian Process Classification (GPC) models, whose kernel structures do not have to be pre-determined, in the sense that both the kernel structures and parameters reside in the latent space of the Bayesian inference.

In this paper, we focus on binary classification, but it would be straightforward to extend our approach to support multi-class classification.

\subsection{Preliminaries}

Let $\mathcal{D} := (\mathbf{X}, \mathbf{y})$ be a dataset for binary classification, where $\mathbf{X} := [\mathbf{x}_1,\ldots,\mathbf{x}_n]$ is the $n$ data points and
$\mathbf{y} := [y_1,\ldots,y_n]$ is the corresponding labels, i.e., $\mathbf{x}_i \in \mathcal{X}$ and $y_i \in \{0,1\}$ for each $i \in [n]$, where $\mathcal{X}$ stands for the feature space of data points.
GPC typically samples a function $f : \mathcal{X} \to \mathbb{R}$ from a Gaussian Process prior $f \sim \mathrm{GP}(m,k)$, where $m : \mathcal{X} \to \mathbb{R}$ is a mean function and $k: \mathcal{X} \times \mathcal{X} \to \mathbb{R}$ is a covariance function, i.e., a \emph{kernel}~\cite{seeger2004gaussian}.
The probabilistic model for classification $P_\mathbf{X}(\mathbf{y}; m,k)$ is then be formulated by
\begin{align}
    [f(\mathbf{x}_1),\ldots,f(\mathbf{x}_n)] & \sim \mathrm{MultivariateNormal}(m(\mathbf{X}), k(\mathbf{X},\mathbf{X})), \label{vgpc:latent} \\
    y_i & \sim \mathrm{Bernoulli}(\sigma(f(\mathbf{x}_i))), \quad i \in [n], \label{vgpc:bern}
\end{align}
where $\sigma : \mathbb{R} \to [0,1]$ is a \emph{sigmoid} function, e.g., the logistic function or the probit function.
To infer the label $y_*$ on a new data point $\mathbf{x}_*$, i.e., to reason about the posterior distribution $P_{\mathbf{X},\mathbf{x}_*}(y_* = 1 \mid \mathbf{y})$, we treat $f(\mathbf{X})$ as a vector $\mathbf{f} := [f_1,\ldots,f_n]$ of latent variables and carry out the inference in two steps: (i) first derive the distribution of the latent variable $f_*$ as
\begin{equation}\label{infer:latent}
P_{\mathbf{X},\mathbf{x}_*}(f_* | k, \mathbf{y}) = \int_{\mathbb{R}^n} P_{\mathbf{X},\mathbf{x}_*}(f_* \mid k, \mathbf{f}) P_{\mathbf{X}}(\mathbf{f} \mid k, \mathbf{y}) d\mathbf{f},
\end{equation}
and (ii) then derive the posterior of $y_*$ by integrating out $f_*$ as 
\begin{equation}\label{infer:label}
P_{\mathbf{X},\mathbf{x}_*}(y_* = 1 | k,\mathbf{y}) = \int_{\mathbb{R}} \sigma(f_*) P_{\mathbf{X},\mathbf{x}_*}(f_* \mid k, \mathbf{y}) d f_*.
\end{equation}
Note that the key part is to account for the posterior distribution of latent variables $\mathbf{P}_{\mathbf{X}}(\mathbf{f} \mid k, \mathbf{y})$.

Usually, the mean $m$ is pre-determined to be the constant-zero function, whereas the kernel $k$ is parameterized by a vector $\theta = [\theta_1,\ldots,\theta_{d(k)}] \in \Theta_k \subseteq \mathbb{R}^{d(k)}$, where $d(k)$ is the number of real-valued parameters in $k$. 
Let us denote $k_\theta := (k, \theta)$ and reuse the same notation for the actual covariance function derived from $k$ and $\theta$.
While many GP-based methods pre-determine $k$ and treat $\theta$ as hyper-parameters, in this paper, we aim to characterize both $k$ and $\theta$ as latent information of the classification model, as well as develop a method that is adaptive in both the structure and parameters of the kernel in the following sections.

\subsection{GPC with A Domain Specific Language for Kernels}

To allow a rich prior distribution over kernel structures $k$, we define a sample space $\mathcal{K}$
using a probabilistic context-free grammar (PCFG), following the practice of a line of prior work~\cite{RN50,10.1145/3290350,saad2023sequential}:

\begin{align}
B \quad &::= \quad \textsc{Linear}\quad | \quad \textsc{SquaredExp}\quad |\quad \textsc{GammaExp} \quad | \quad \ldots  \label{kernel:base} \\
\oplus \quad &::= \quad + \quad | \quad \times \quad \label{kernel:combine} \\
k \quad &::= \quad B \quad | \quad (k_1 \oplus k_2)  \label{kernel:kernel}
\end{align}
The non-terminal $B$ stands for an extensible collection of  basic kernels.
Two kernels can be combined with a binary operator $\oplus$ that computes the pointwise addition or multiplication of the two kernels.
The PCFG also assigns a probability to each production rule in \eqref{kernel:base}-\eqref{kernel:kernel}, thus formulating a prior distribution on $\mathcal{K}$.
The meaning of a kernel expression $k \in \mathcal{K}$ is defined inductively as follows.
\begin{align}
\textsc{Linear}_{\alpha, \beta, \mathbf{w}}(\mathbf{x}, \mathbf{x}')  & :=  \alpha + (\mathbf{x} - \mathbf{w})^{\top}(\mathbf{x}' - \mathbf{w}), && \alpha >0  \\
\textsc{SquaredExp}_{\ell}(\mathbf{x}, \mathbf{x}')  & := \exp (-\lVert \mathbf{x} - \mathbf{x}'\rVert ^2/{2 \ell^2}), && \ell > 0  \\
\textsc{GammaExp}_{\ell,\gamma}(\mathbf{x}, \mathbf{x}')  & := \exp (-(\lVert \mathbf{x} - \mathbf{x}'\rVert/\ell)^\gamma ), && \ell > 0, 0 < \gamma < 2 \\
(k_1 \oplus k_2)_{\theta_1,\theta_2}(\mathbf{x},\mathbf{x}') & := k_1{}_{\theta_1}(\mathbf{x},\mathbf{x}') \oplus k_2{}_{\theta_2}(\mathbf{x},\mathbf{x}')
\end{align}

With the PCFG, we extend the standard GPC model (c.f. \eqref{vgpc:latent}-\eqref{vgpc:bern}) to treat both the structure $k$ and the parameters $\theta$ of the kernel as latent variables, resulting in a probabilistic model $P_{\mathbf{X}}(k,\theta,\epsilon,\mathbf{y})$:
\begin{align}
k  &\sim  \textrm{PCFG, c.f.(\ref{kernel:base}) - (\ref{kernel:kernel})} \label{gpc:k} \\
[\theta_1, \ldots, \theta_{d(k)}]  &  \overset{\mathrm{iid}}{\sim}  \mathrm{Normal} (0, 1)  \\
\epsilon  &\sim  \mathrm{InverseGamma} (0, 1)   \\
[f_1,\ldots,f_n] & \sim \mathrm{MultivariateNormal}(\mathbf{0}, k_\theta(\mathbf{X},\mathbf{X}) + \epsilon \cdot \mathbf{I}) \label{gpc:f} \\
y_i & \sim \mathrm{Bernoulli}(\sigma(f_i)), \quad i \in [n] \label{gpc:label}
\end{align}
The latent variable
$\epsilon$ stands for the noise in the GP.
Note that the model samples kernel parameters from the standard Normal distribution, but basic kernels may impose constraints on its parameters.
We apply standard transformations to obtain constrained parameters and omit the details here; for example, $z \mapsto \exp(z)$ for positive parameters
and $z \mapsto 2/(1+\exp(z))$ for parameters in $(0,2)$.

We further apply a standard \emph{reparameterization} trick to the model above via Cholesky decomposition.
That is, instead of sampling from a multivariate Normal distribution, 
we sample $n+1$ i.i.d. auxiliary variables from the standard Normal distribution
and compute the latent vector $\mathbf{f}$ as follows.
\begin{align}
\beta  &\sim \mathrm{Normal} (0, 1)   \\
\boldsymbol{\eta} := [\eta_1,\ldots,\eta_n]  &\overset{\mathrm{iid}}{\sim}  \mathrm{Normal} (0, 1)   \\
\mathbf{L}  &=  \mathrm{Cholesky}(k_\theta(\mathbf{X},\mathbf{X}) + \epsilon \cdot \mathbf{I})  \\
\mathbf{f}  &=  \mathbf{L} \cdot \boldsymbol{\eta} + \beta 
\label{getf}
\end{align}
The term $\mathrm{Cholesky}(\mathbf{M})$ performs Cholesky decomposition of the covariance matrix $\mathbf{M}$, i.e., computes a lower-triangular matrix $\mathbf{L}$ such that $\mathbf{L} \mathbf{L}^\top = \mathbf{M}$.
The reparmeterized model specifies a joint distribution $P_\mathbf{X}(k,\theta,\epsilon,\beta,\boldsymbol{\eta})$.
To simplify the notation, we define $\phi := (\theta,\epsilon)$
and $\psi := (\beta,\boldsymbol{\eta})$ to denote the kernel parameters and auxiliary variables, respectively.

\subsection{Problem Statement: Structure Discovery for GPC}

By the Bayes' rule, the posterior distribution on $k,\phi,\psi$ given a dataset $(\mathbf{X},\mathbf{y})$ is given by
\begin{equation} \label{ps:post}
P_\mathbf{X}(k,\phi,\psi \mid \mathbf{y}) = \frac{ P_\mathbf{X}(k,\phi,\psi, \mathbf{y})}{ P_\mathbf{X}(\mathbf{y}) } = \frac{P_{\mathbf{X}}(k,\phi, \psi,\mathbf{y})}{\sum_{k \in \mathcal{K}}\int_{\Theta_k \times \mathbb{R}_+ \times \mathbb{R} \times \mathbb{R}^n}P_{\mathbf{X}}(k,\phi, \psi,\mathbf{y}) d(\phi,\psi) }.
\end{equation}

The goal of our method is to generate and maintain a finite set of $M \ge 1$ weighted \emph{particles}
\begin{align*}
\{(w^i, (k^i, \phi^i, \psi^i))\}_{i=1}^{M}, \numberthis
\label{weight}
\end{align*}
each of which consists a weight $w^i > 0$ and a tuple of latent variables $(k^i,\phi^i,\psi^i)$.
These particles are intended to approximate the posterior distribution given in \eqref{ps:post}, so as to compute the expectation of some test function $\tau$ with respect to the posterior distribution as

\begin{align*}
\label{expect}
\mathbb{E}_{P_{\mathbf{X}}}[\tau(k,\phi,\psi,\mathbf{X},\mathbf{y}) \mid \mathbf{y}] \approx \sum\limits_{i=1}^{M}\frac{w^i}{\sum_{j}w^j}\tau(k^i,\phi^i,\psi^i,\mathbf{X},\mathbf{y}). \numberthis
\end{align*}

Recall that in \eqref{infer:latent}-\eqref{infer:label} we reviewed how to condition a standard GPC model on a dataset to make predictions.
We generalize the idea to define a test function $\tau^\mathsf{prob}_{\mathbf{x}_*}$ to compute the classification probability of a new data point $\mathbf{x}_*$, given all the latent information $k,\phi,\psi$:
\begin{equation}\label{test:prob}
    \tau^\mathsf{prob}_{\mathbf{x}_*}(k,\phi,\psi,\mathbf{X},\mathbf{y}) := \int_{\mathbb{R}} \sigma(f_*) P_{\mathbf{X},\mathbf{x}_*}(f_* \mid k,\phi,\psi,\mathbf{y}) d f_* .
\end{equation}
Different from \eqref{infer:latent}, the posterior distribution on $f_*$ becomes analytically tractable because the latent vector $\mathbf{f}$ is determined by the given latent information $k,\phi,\psi$.
To see that, we derive  and simplify the posterior distribution on $f_*$ as
\begin{align}
P_{\mathbf{X},\mathbf{x}_*}(f_* \mid \mathbf{f}, \mathbf{y})  =
 \frac{P_{\mathbf{X},\mathbf{x}_*}(f_*,\mathbf{f},\mathbf{y})}{P_\mathbf{X}(\mathbf{f},\mathbf{y}) } 
 = \frac{P_{\mathbf{X},\mathbf{x}_*}(f_*, \mathbf{f}) P_\mathbf{X}(\mathbf{y} \mid \mathbf{f}) }{ P_\mathbf{X}(\mathbf{f}) P_\mathbf{X}(\mathbf{y} \mid \mathbf{f}) }
 = P_{\mathbf{X},\mathbf{x}_*}(f_* \mid \mathbf{f}) , \label{latent}
\end{align}
which is a Normal distribution because of the multivariate Normal joint distribution (c.f. \eqref{gpc:f}).
However, \eqref{test:prob} might not be analytically tractable due to the choice of the sigmoid function $\sigma$.
Fortunately, to predict the label for a new data point $\mathbf{x}_*$, \eqref{test:prob} is essentially a uni-variate integral, so we resort to use Monte Carlo estimation.
In some case, e.g., when $\sigma$ is the probit function, the integral has a closed-form solution so that $\tau^\mathsf{prob}_{\mathbf{x}_*}$ can be evaluated easily.

\section{Sequential Monte Carlo for Adaptive and Incremental GPC Learning}
\label{Online}

In this section, we develop an adaptive and incremental learning method for the GPC models present in Section \ref{automatic}.
We extend a recently proposed Sequential Monte Carlo sampler for time series learning, namely AutoGP~\cite{saad2023sequential}, to our setting of GP-based classification.
Notably, our method supports both online and offline settings, enabling itself to perform classification both when data arrive sequentially and when data are available all at once.

\subsection{Background}

\paragraph{Sequential Monte Carlo}
Sequential Monte Carlo (SMC) is a class of sampling-based inference algorithms designed to approximate a sequence $\pi_0,\pi_1,\ldots,\pi_T$ of hard-to-sample probability distributions, especially in dynamic and non-linear systems~\cite{smcmcmc}.
An SMC sampler produces at each step $j=0,1,\ldots,T$ a finite set of weighted particles $\{(w^i_j,x^i_j)\}_{i=1}^M$---in the same manner as shown by \eqref{weight}---as an empirical approximation of the distribution $\pi_j$.
Initially at step $0$, an SMC sampler draws i.i.d. samples from $\pi_0$ (assuming that $\pi_0$ is easy to sample instead) and assigns all the weights to be one.
At step $j$, the particle set $\{(w^i_{j-1},x^i_{j-1})\}_{i=1}^M$ is updated in two steps: (i) first evolve each particle by sampling from a \emph{forward} Markov kernel $K$ between the measurable spaces at step $j-1$ and $j$:
\begin{equation} 
x_j^i \sim K_j(x_{j-1}^i, \cdot), 
\end{equation}
and (ii) then reweight each particle using the forward Markov kernel $K_j$ as well as a \emph{backward} Markov kernel $L_{j-1}$ between the measurable spaces at step $j$ and $j-1$:
\begin{equation} 
w_j^i \gets w_{j-1}^i \cdot  \left(  \frac{\pi_j(x_j^i) L_{j-1}(x_j^i,x_{j-1}^i)}{\pi_{j-1}(x_{j-1}^i)K_{j}(x_{j-1}^i,x_{j}^i)}  \right),
\end{equation} 
with the understanding that it is totally fine if we can evaluate $\pi_j(\cdot)$ pointwise up to a normalizing factor.
Those Markov kernels should be chosen based on the actual learning problem.
For example, one can define $L$ to be the time reversal of $K$ such that $\pi_j(x_j) L_{j-1}(x_{j}, x_{j-1}) = \pi_{j}(x_{j-1}) K_j(x_{j-1},x_j)$, leading to a simple reweight scheme $w_j^i \gets w_{j-1}^i \cdot \frac{ \pi_j(x_{j-1}^i) }{ \pi_{j-1}(x_{j-1}^i) }$.
Note that an SMC sampler uses the backward kernels only to do density estimation, but it uses the forward kernels also for proposing new values.
An SMC sampler also features a \emph{resampling} phase that deals with particle collapse, i.e., when the weights of some particles become negligible compared against other particles.
Each particle $x_j^i$ would be resampled, simultaneously, to use the value $x_j^{i'}$ with probability $w_j^{i'}/(w_j^1 + \ldots + w_j^M)$.
After resampling, an SMC sampler can use a \emph{rejuvenation} phase to evolve the particles within the step $j$, i.e., to rejuvenate each particle with respect to the target distribution $\pi_j.$
There are many ways to implement rejuvenation; in this paper, we consider using Markov-Chain Monte Carlo (MCMC).

\paragraph{Markov Chain Monte Carlo}
Markov Chain Monte Carlo (MCMC) methods are a class of inference algorithms that sample from hard-to-sample probability distributions by constructing a Markov chain, which has the desired target distribution as its stationary distribution.
An MCMC method generates a sequence of samples by simulating the Markov chain, so that every sample only depends on the previous sample.
At the heart of MCMC methods is the Metropolis-Hastings (MH) algorithm  \cite{hastings1970monte, metropolis1953equation}.
MH starts to iterate by proposing $x'$ from a proposal distribution $q(x';x)$, where $q(x';x)$ also denotes the probability density that previous state $x$ transits to the proposed state $x'$.
MH then computes a ratio to decide whether to accept $x'$:
\begin{equation}
\alpha(x', x) = \min\left( 1, \frac{\pi(x')q(x;x')}{\pi(x)q(x';x)} \right),
\end{equation}
where $\pi(\cdot)$ is the target distribution, which can be evaluated up to a normalizing constant.
Two popular and powerful MCMC methods are Involutive MCMC (IMCMC)~\cite{ICML:NWE20} and Hamiltonian Monte Carlo (HMC)~\cite{kn:Neal10}.
IMCMC's proposal samples auxiliary variables $y \in \mathcal{Y}$ and uses an involutive map $f$ on $\mathcal{X} \times \mathcal{Y}$ (i.e., $f=f^{-1}$) to propose a new state $(x',y') = f(x,y)$, where $x \in \mathcal{X}$ is the current state.
As we will soon discuss in the next paragraph, IMCMC is suitable to implement \emph{transdimensional} proposals; in particular, it is suitable to evolve the kernel structure in GP-based models.
HMC takes inspiration from Hamiltonian dynamics, and its proposal samples an auxiliary momentum variable and uses numerical integration (e.g., leap-frog) to generate diverse candidate states in a continuous sample space.
In particular, it is suitable to evolve the real-valued random variables in our GPC model, including the kernel parameters $\phi$ and auxiliary variables $\psi$.

\paragraph{SMC for Time Series Learning}
AutoGP~\cite{saad2023sequential} is a recently proposed method that is able to effectively and adaptively find suitable structures for time-series data in both the offline and online settings.
Notably, AutoGP uses Gaussian Process Regression for time-series learning.
In this paper, we take inspiration from AutoGP and extend its methodology to solve classification problems.
AutoGP is an SMC sampler based on \emph{data tempering}; that is, the target distribution $\pi_j$ at step $j$ is the posterior distribution conditioned on the first $j$ data points in the time-series data.
In this way, AutoGP enables \emph{incremental} learning, and the nature of SMC---which evolves a collection of particles with possibly different kernel structures---enables \emph{adaptive} learning.
AutoGP carefully designs an IMCMC proposal to rejuvenate kernel structures.
The proposal makes use of \textsc{Subtree-Replace} operations and \textsc{Detach-Attach} operations.
The \textsc{Subtree-Replace} operation randomly replaces a sub-structure of a given kernel to another one, while the \textsc{Detach-Attach} operation randomly moves a sub-structure of a given kernel to another location in the same kernel.
These operations enable AutoGP to effectively and efficiently explore versatile kernel structures.

\subsection{Method}

Algorithm \ref{algo} shows our method of applying SMC to adaptive and incremental learning on the GPC models present in Section \ref{automatic}.
It follows a standard reweight-resample-rejuvenate pipeline for implementing SMC samplers~\cite{step}. 
Our method reweights and resamples the particle set $\{(k^i_j, \phi^i,\psi^i_j)\}_{i=1}^M$ as shown in line \ref{line1} to line \ref{line2}.
Line \ref{line1} and line \ref{lineWe} are the process of reweighting;
it implicitly chooses the forward Markov kernel $K$ to be the identity kernel
and the backward Markov kernel $L$ to be its time reversal.
As a result,
it calculates the new weight for each particle based on the joint probability of model's latent variables $(k,\phi,\psi)$ and the first $j$ data points $(\mathbf{X}_{1:j}, \mathbf{y}_{1:j})$.
The process of resampling in lines \ref{lineSa} to \ref{line2} adopts a standard machinery of \emph{adaptive} resampling, i.e., only initiate the resample process when the effective sample size (ESS) drops below a threshold. 
Finally, lines \ref{reju} to \ref{reju:end} implement the rejuvenation loop.
In each rejuvenation iteration, we follow the design of AutoGP~\cite{saad2023sequential} to first evolve the kernel structure of each particle via IMCMC (reviewed in the previous section) and then apply HMC to evolve the real-valued latent kernel parameters $\phi$ and auxiliary variables $\psi$.

Algorithm \ref{algo} can already to applied to the online setting, in the sense that the step $j$ stands for the order when a data point arrives.
In the offline setting, although algorithm \ref{algo} can also work, we can make it more effective by allowing \emph{batch tempering}, i.e., we can divide a dataset $\mathcal{D} = (\mathbf{X},\mathbf{y})$ into $T$ batches, each of which contains $n/T$ data points $\mathcal{D}_j = (\mathbf{X}_{j\cdot n/T+1:(j+1) \cdot n/T}, \mathbf{y}_{j\cdot n/T+1 : (j+1) \cdot n/T})$, for $j = 0, \ldots, T-1$.
Then in the iteration for step $j$, instead of incorporating one single data point, we use the whole batch $\mathcal{D}_j$ in the reweighting and rejuvenation processes.

With the particle set $\{(w^i,(k^i,\phi^i,\psi^i))\}_{i=1}^M$ that approximates the posterior distribution $P_\mathbf{X}(k,\phi,\psi \mid \mathbf{y})$, we can use it to make predictions for new data points.
Following \eqref{expect}-\eqref{latent}, for a new data point $\mathbf{x}_*$,
we approximately compute the predictive probability of $y_*$ being $1$ as
\begin{equation}
    \sum_{i=1}^M \frac{w^i}{\sum_j w^j} \frac{1}{K} \sum_{k=1}^K \sigma(f_k^i), 
\end{equation}
where $K$ is the number of Monte Carlo estimations to approximate $\tau^\mathsf{prob}_{\mathbf{x}_*}$ and $f^i_1,\ldots,f^i_K$ are $K$ i.i.d. samples from the posterior $P_{\mathbf{X},\mathbf{x}_*}(\cdot \mid \mathbf{f}^i)$ for each particle $i$.

\begin{figure}[t]
  \centering
  \begin{minipage}{.9\linewidth}
    \begin{algorithm}[H]
      \SetAlgoLined
      \KwData{classification dataset $(\mathbf{X},\mathbf{y})$ of $n$ data points}
      \KwIn{number of particles $M > 0$ and number of rejuvenation steps $N_{reju} > 0$}
      \KwResult{weighted samples $(w^i, (k^i, \phi^i, \psi^i))$ for posterior distribution $P_{\mathbf{X}}(k,\phi,\psi \mid \mathbf{y})$} 
      \tcp{Repeat operations on particles indexed by $i$ over $i=1,\ldots, M$}
      $(k^{i}_{0},\phi^{i}_{0},\psi^i_0) \sim P(k, \phi, \psi)$\;
      $w_{0}^{i}\gets 1$\;
      \For {$j=1,\ldots, n$}{
        $(k^{i}_{j},\phi^{i}_{j},\psi^i_j) \gets (k^{i}_{j-1},\phi^{i}_{j-1}, \psi^i_{j-1})$\;
        \label{line1}
        $w_{j}^{i}\gets w_{j-1}^{i} \cdot \Bigl(\frac{P_{\mathbf{X}_{1:j}}\;(k^i_j,\phi^i_j, \psi^i_j, \mathbf{y}_{1:j})}{P_{\mathbf{X}_{1:j-1}}(k^i_{j-1},\phi^i_{j-1}, \psi^i_{j-1},\mathbf{y}_{1:j-1})}\Bigr)$ \;
        \label{lineWe}
        \If{$j < n$ and \textbf{RESAMPLE?}($w_{j}^{1:M}$)}{
        \label{lineSa}
          $(l_1,\dots,l_M) \gets $\textbf{\textit{RESAMPLE}}($w_{j}^{1:M}$)\;
          $(k^{i}_{j},\phi^{i}_{j},\psi^i_j) \gets (k^{l_i}_{j-1},\phi^{l_i}_{j-1},\psi^{l_i}_{j-1})$\;
          $w_{j}^{i}\gets (w_{j}^{1}+\dots+w_{j}^{M})/M$ 
        }
        \label{line2}

        \For{$u = 1,\ldots,N_{reju}$}{
        \label{reju}
        $(k^{i}_{j},\phi^{i}_{j}) \sim   \textbf{\textit{IMCMC}}((k, \phi) \mapsto P_{\mathbf{X}_{1:j}}(k,\phi, \psi^i_j ,\mathbf{y}_{1:j});(k^{i}_{j},\phi^{i}_{j}))$\; \label{imcmc}
        $(\phi^{i}_{j},\psi^i_j) \sim \textbf{\textit{HMC}}((\phi,\psi) \mapsto P_{\mathbf{X}_{1:j}}(k_{j}^{i},\phi, \psi,\mathbf{y}_{1:j}); \phi^{i}_{j}, \psi^i_j)$
        \label{hmc}
        }
        \label{reju:end}
      }
      \caption{SMC Learning for Classification}
      \label{algo}
    \end{algorithm}
  \end{minipage}
\end{figure}

\section{Experiment}
\label{experiment}
We implemented our method based on AutoGP~\cite{saad2023sequential}.
We set up our experiments on a device with 13th Gen Intel(R) Core(TM) i9-13900H 2.60 GHz CPU and 32 GB RAM.  

The research questions of the experiments are to study the following: 
\begin{itemize}
    \item \textbf{RQ1}: Can our method learn kernel structures and parameters \emph{adaptively} for classification?
    \label{aim1}
    \item \textbf{RQ2}: Can our method learn kernel structures and parameters \emph{incrementally} for classification?
    \label{aim2}
\end{itemize}
For the purpose of testing our method comprehensively, there are two parts in our experiment.
The first part involves applying our method to some toy datasets to demonstrate the capability and characterization of our method.
The second part involves applying our method to real-world datasets across diverse domains, assessing its performance under authentic data conditions. 
 
\subsection{Toy Datasets}
\label{toydata}
\begin{figure}[t]
  \centering
    \includegraphics[width=.8\textwidth]{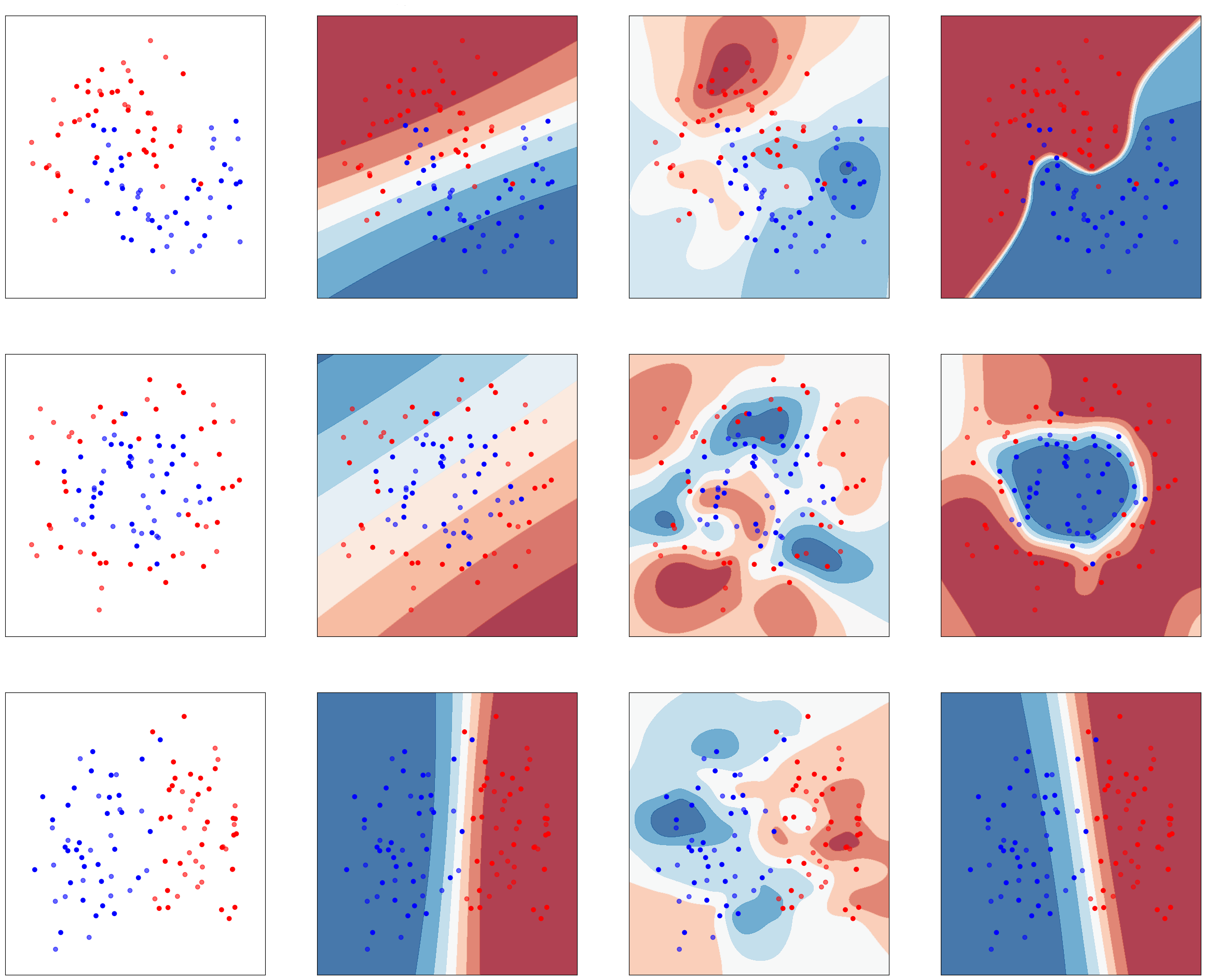}
    \caption{Contour plots of our method with \textsc{Linear} and \textsc{SquaredExponential}  GP. From left to right: original data points, the result of \textsc{Linear} GP, the result of \textsc{SquaredExponential} GP, and the result of our method. It shows that our method can discover different kernel structures.} 
  \label{fig1}

\end{figure}
The toy datasets used in our experiments are generated by Scikit-Learn \cite{sklearn}.
Those toy datasets with 2-dimensional inputs are easy to be visualized with weights; thus, they provide an intuitive way to illustrate the characterization of our method.

For \textbf{RQ1}, the first experiment is designed to demonstrate the automatic adaptability of our method in selecting appropriate kernels for varying datasets. Figure \ref{fig1} illustrates the comparison between our method and GPs using pre-selected kernels---\textsc{Linear} and \textsc{SquaredExponential}---across three datasets. GPs used in this experiment are implemented using the same machinery as Algorithm \ref{algo}, except that the kernel structure $k$ is fixed.
This comparison demonstrates that our method can exhibit characteristics of different kernels. 

\begin{figure}[htb]
\centering
\begin{subfigure}{0.2\textwidth}
    \includegraphics[width=\textwidth]{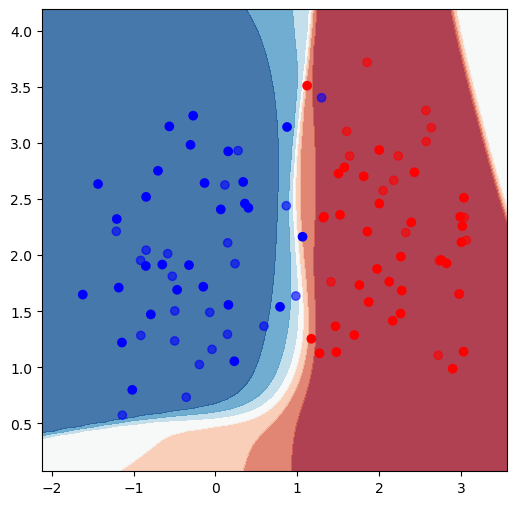}
    \caption{}
    \label{fig:first}
\end{subfigure}
\hfill
\begin{subfigure}{0.2\textwidth}
    \includegraphics[width=\textwidth]{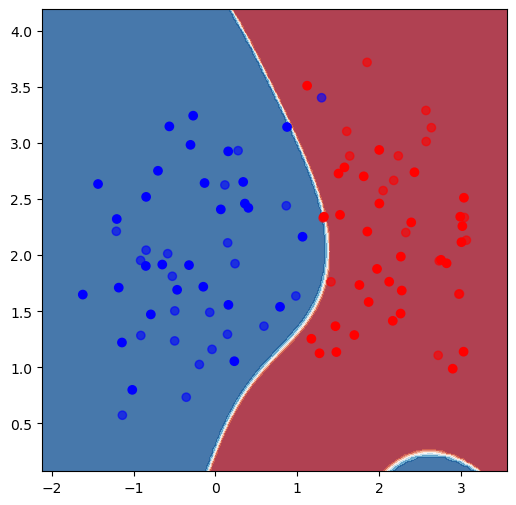}
    \caption{}
    \label{fig:second}
\end{subfigure}
\hfill
\begin{subfigure}{0.2\textwidth}
    \includegraphics[width=\textwidth]{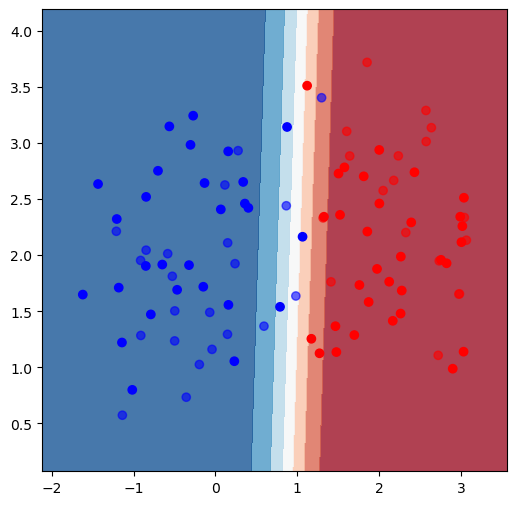}
    \caption{}
    \label{fig:third}
\end{subfigure}
 \hfill
\begin{subfigure}{0.2\textwidth}
    \includegraphics[width=\textwidth]{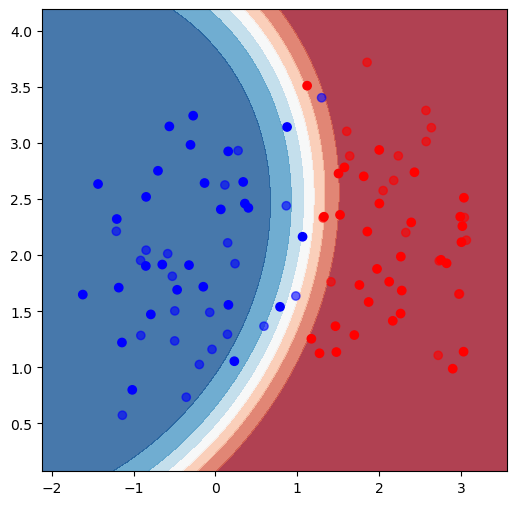}
    \caption{}
    \label{fig:four}
\end{subfigure}

\caption{Contour plots of different particles after learning a linearly separable dataset. Fig.\ref{fig:first} is the result with all particles. Fig.\ref{fig:second} is the particle with kernel $\textsc{SquaredExponential} \times (\textsc{Linear} + \textsc{SquaredExponential})$. Fig.\ref{fig:third} is the particle with kernel \textsc{Linear}. Fig.\ref{fig:four} is the particle with kernel $\textsc{Linear} \times \textsc{Linear}$. }
\label{fig:figures}
\end{figure}

To take a closer look at on how our method combines properties of different kernels, Figure \ref{fig:figures} plots the kernel's behavior of different particles after running Algorithm \ref{algo} on a dataset that can be linearly separated.
It shows that our method could discover different kernel structures with similar performance; thus, the method would be robust to incorporate future unseen data points.

\begin{figure}[t]
\centering

\begin{subfigure}[b]{.4\textwidth}
\centering
    \includegraphics[width=.8\textwidth]{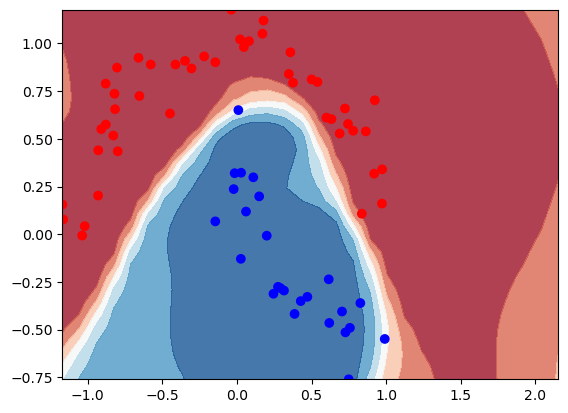}

\caption{Result with First Batch}
\end{subfigure}\quad
\begin{subfigure}[b]{.4\textwidth}
\centering
    \includegraphics[width=.8\textwidth]{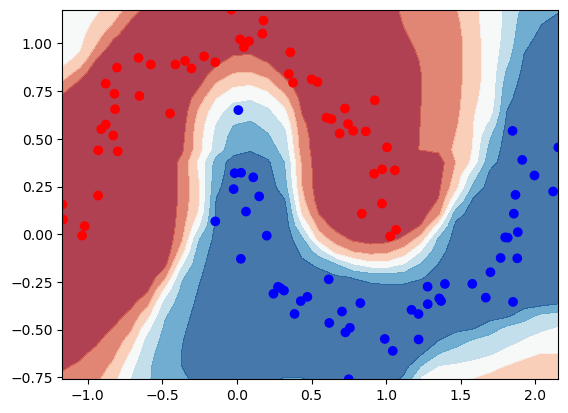}

\caption{Result by adding rest of data}
\end{subfigure}

\caption{Contour plots of our method in the context of patterns shifts. The left figure is the result of initial learning. The right figure is the result of incorporating another batch with the remaining data.}
\label{shift}
\end{figure}

Furthermore, to evaluate our method's performance on pattern shifts and simulate an online setting described in \textbf{RQ2}, we initially run our method to learn on a portion of the dataset, followed by subsequently learning on the remaining data in another batch.
Figure \ref{shift} shows that our method is able to adjust kernel structures to adapt to the shift of pattern in an online setting.

In summary, experiments on the toy datasets show that our method can discover various kernel structures, as well as learn and adapt to pattern shifts in the online setting.

\subsection{Real-World Datasets}

To test our method in practice, we select three real-world datasets in different domain. Some datasets \cite{musk, sigillito1989} were used in previous studies \cite{lu2022incremental}. The details of datasets can be found at Appendix.

We use two GPs with pre-determined kernels, Random Forest, and three other classification methods for evaluating \textbf{RQ1}.
The two GPs are set up in the same manner as Section \ref{toydata}.
We use a Julia implementation of Random Forest \cite{ben_sadeghi_2022_7359268}.
To further test the effectiveness of our method, we select three methods provided by Scikit-Learn \cite{sklearn}.
We choose Passive Aggressive Classifier (PAC) \cite{pac}, Stochastic Gradient Descent (SGD) \cite{sdg}, and Naive Bayes in this experiment.

In order to test the ability of online learning for \textbf{RQ2}, we adapt a similar setting to Section \ref{toydata} but make it more extreme and biased:
in the first batch of learning, we only include one class of the data
and then incorporate batches of remaining data with the other class.
%

Results are shown in Table \ref{res}.
Accuracy of each offline task is calculated in a standard way and reported.
For online task, average accuracy of each task is calculated by $\frac{1}{B}\sum^{B}_{i=1} A_i$, where $B$ is the total number of batches and $A_i$ is accuracy after each batch of learning. 
For the offline setting, our method outperforms all other methods. It shows that our method achieves the goal of reducing requirement of prior knowledge by automatic selection of kernel.
For the online setting, our method achieves highest accuracy on two out of three datasets. The results of the online setting demonstrate our method is able to adapt kernel structures in an incremental manner.

In summary, experiments on the real-world datasets show that our method has ability to seamlessly integrate features of different kernels and accurately adjust itself against pattern shifts within the dataset.

\begin{table}[bt]
\centering
\caption{Experimental results of real-world datasets.}
\begin{tabular}{c c c c c c c c c}
\hline

&\multicolumn{3}{c}{Offline} &\multicolumn{3}{c}{Online}\\
\cmidrule(r){2-4} \cmidrule(r){5-7}
                    &Ionosphere  & Musk   &Heartdisease  &Ionosphere   & Musk  &Heartdisease   \\  \hline
Our Method          &  \textbf{90.7\%} &\textbf{86.3\%} & \textbf{89.9\%} & \textbf{81.1\%} & \textbf{61.7\%} & 72.0\%\\ 
Linear &  87.2\% & 61.7\% & 82.3\% & 74.4\% & 54.9\% & 66.7\%\\
SquaredExponential &  84.3\% & 65.4\% & 84.0\% & 73.2\% & 58.9\% & 70.7\%\\ 
Naive Bayes         &  75.1\% & 75.3\%  & 84.0\% & 72.9\% & 58.8\% & \textbf{74.6\%} \\ 
PAC  & 80.8\%  & 64.3\% & 80.6\% & 65.6\%  & 56.4\% & 66.6\% \\ 

SGD &  81.5\% & 78.5\% & 76.4\% & 61.8\% & 55.8\% & 71.4\%\\ 
Random Forest &  87.2\% & 76.4\% & 81.5\% & 73.3\% & 52.1\% & 62.2\%\\ 
\hline
\end{tabular}
\label{res}
\end{table}


\section{Related Work}

\paragraph{Automatic Selection of GP Kernels}
Automatic selection of kernels \cite{RN50, 10.1145/3290350,saad2023sequential} is a technique that automatically samples GP kernels, defined within a context-free grammar (CFG), to model various types of data. Earlier works have primarily focused on employing CFG for the regression of data. Specifically, \cite{10.1145/3290350} and \cite{saad2023sequential} have advanced the inference algorithms for automatic selection of kernels.

Our work is based on automatic selection of kernel for regression, namely AutoGP~\cite{saad2023sequential}, with expansion of non-time-series, non-scalar inputs, and classification.

Given the challenge of non-Gaussian posteriors in classification scenarios, we introduce adapt AutoGP to approximate and predict with such posteriors effectively.

\paragraph{Incremental Learning with GPs}
There are many previous works on using GPs to build classification model \cite{lu2022incremental,shen2019random, bui2017streaming}.
Work \cite{lu2022incremental} presents an ensemble learning method that learns an ensemble of GPs and handles incremental data by applying kernel dictionaries that contain pre-selected kernel structures.
Compared with those methods, our method follows the methodology of Bayesian inference to approximate the posterior distribution, thus requiring less prior knowledge when designing the learning algorithm.
By sampling from a CFG space, our method provides more flexibility to build GP-based classification models. However, our method is typically more computationally expensive than non-Bayesian methods.

\paragraph{Sequential Monte Carlo Learning}
SMC learning has a broad field of applications including graphical models \cite{paige2016inference}, finance \cite{dai2022invitation}, and robotic \cite{liang2008information}.
Despite its extensive application in time-related domains, our work seeks to expand the utility of SMC to non-time-related problem, such as classification tasks.
The connection is established via a broadly applied technique called \emph{data tempering}, i.e., organizing the dataset as a sequence of data points or a sequence of batches.
However, data tempering is usually more computationally expensive, especially in our context of GP-based learning, as GP computation is also expensive.

\section{Conclusion}

In this paper, we propose a new method for Gaussian Process classification by  sequential Monte Carlo.
We extend the problem formulation of structure discovery for regression to the binary-classification problem.
Based on a recently proposed method on sequential Monte Carlo for time-series learning, we develop an algorithm for adaptive and incremental learning of both kernel structures and parameters.
Especially, our algorithm is able to handle the non-Gaussian posterior distributions that arise from Gaussian Process classification problems.
The experiments shows that our method can discover different kernel structures for different datasets and outperform various classification methods on real-world datasets. 


\printbibliography

\end{document}